\documentclass{article}
\usepackage{tabularx}
\usepackage{multirow}

\usepackage{arxiv}
\usepackage[utf8]{inputenc} 
\usepackage[T1]{fontenc}    
\usepackage{hyperref}       
\usepackage{url}            
\usepackage{booktabs}       
\usepackage{amsfonts}       
\usepackage{nicefrac}       
\usepackage{microtype}      
\usepackage{lipsum}
\usepackage[bottom]{footmisc} 
\usepackage{graphicx}
\usepackage{amsmath}
\usepackage{array}
\usepackage{wasysym}
\usepackage{multirow}
\usepackage{adjustbox}
\usepackage{authblk}  
\title{Can Multi-modal (reasoning) LLMs detect document manipulation?}
\author[1]{Zisheng Liang\textsuperscript{*}}
\author[2]{Kidus Zewde\textsuperscript{*}}
\author[3]{Rudra Pratap Singh \textsuperscript{*}}
\author[3]{Disha Patil \textsuperscript{*}}
\author[4]{Zexi Chen \textsuperscript{*}}
\author[5]{Jiayu Xue \textsuperscript{*}}
\author[6]{Yao Yao\textsuperscript{*}}
\author[7]{Yifei Chen\textsuperscript{*}} 
\author[8]{Qinzhe Liu\textsuperscript{*}}
\author[2]{Simiao Ren\textsuperscript{\dag}}

\affil[1]{Duke University, \texttt{zisheng.liang@duke.edu}}
\affil[2]{Scam.ai, \texttt{kidus.zewde/benren@scam.ai}}
\affil[3]{Indian Institute of Technology, Roorkee, \texttt{p\_dchetan/r\_psingh@ma.iitr.ac.in}}
\affil[4]{New York University, \texttt{zc2610@nyu.edu}}
\affil[6]{University of Wisconsin Madison, \texttt{yyao39/xzhan23@wisc.edu}}
\affil[5]{University of North Carolina at Chapel Hill \texttt{xuejiayu@ad.unc.edu}}
\affil[7]{Columnbia University, \texttt{yc3503@columbia.edu}}
\affil[8]{\texttt{qinzhe1126@gmail.com}}

\begin{document}
\maketitle
{\renewcommand{\thefootnote}{}\footnote{Authorship of this paper does not imply employment by or affiliation with Scam.AI (Reality Inc.); some authors are independent researchers who collaborated on this work.}\addtocounter{footnote}{-1}}

\begin{abstract}
Document fraud poses a significant threat to industries reliant on secure and verifiable documentation, necessitating robust detection mechanisms. This study investigates the efficacy of state-of-the-art multi-modal large language models (LLMs), including OpenAI O1/4o, Gemini Flash (thinking), Deepseek Janus, Grok, Llama 3.2 / 4, Qwen 2/2.5 VL, Mistral Pixtral, and Claude 3.5/3.7 Sonnet, in detecting fraudulent documents. We benchmark these models against each other and prior work on document fraud detection techniques using a standard dataset with real transactional document. Through prompt optimization and detailed analysis of the models’ reasoning processes, we evaluate their ability to identify subtle indicators of fraud, such as tampered text, misaligned formatting, and inconsistent transactional sum. Our results reveal that top-performing multi-modal LLMs demonstrate superior zero-shot generalization, outperforming conventional methods on out-of-distribution datasets, while several vision LLMs exhibit inconsistent or subpar performance. Notably, model size and advanced reasoning capabilities show limited correlation with detection accuracy, suggesting task-specific fine-tuning is critical. This study underscores the potential of multi-modal LLMs in enhancing document fraud detection systems and provides a foundation for future research into interpretable and scalable fraud mitigation strategies.

\end{abstract}

\section{INTRODUCTION}

\renewcommand{\thefootnote}{\fnsymbol{footnote}}

\footnotetext[2]{\dag\ Correspondence author.}
\footnotetext[1]{* Equal contributions, order randomly generated.}

The emergence of artificial intelligence has yielded significant advancements across various domains, from material design \cite{khatib2021deep}, energy infrastructure \cite{ren2022automated} to deepfake detection \cite{ren2025can}. However, the rapid advancement of generative models has facilitated the creation of highly convincing forged documents, such as receipts, posing serious risks to financial integrity, legal authenticity, and institutional trust. Traditional forgery detection methods often rely on rule-based systems or convolutional neural networks (CNNs) to identify inconsistencies in document features \cite{amerini2011sift, christlein2012evaluation, fridrich2003detection, pan2012splicing, popescu2004exposing, teerakanok2019copy}. Such methods have been extensively surveyed in the literature on image forgery detection \cite{sharma2023comprehensive, zanardelli2023survey}. Yet, as forgery techniques become more sophisticated, these methods struggle to generalize, particularly when encountering novel manipulations not present in training data.

Recent developments in multi-modal large language models (LLMs) have showcased their ability to process and reason across multiple data modalities, including text and images. Models like GPT-4V and Gemini, with their extensive pretraining corpora and emergent reasoning capabilities, are promising tools for detecting forged documents. Unlike traditional methods, multi-modal LLMs can leverage contextual understanding, analyze visual and textual document features, and adapt to previously unseen forgery patterns.

In this study, we systematically evaluate the effectiveness of multi-modal LLMs for detecting forged receipts using the "FindIt Again" dataset, which contains a diverse collection of forged and authentic receipts. We benchmark several state-of-the-art (SOTA) models to assess their generalization capabilities across this dataset. Additionally, we perform ablation studies and probe their reasoning processes to uncover the key factors influencing their decision-making. Our analysis provides insights into how these models reason about forged documents, highlighting their interpretability and limitations.

Our contributions are threefold:

\begin{itemize}
    \item Benchmarking SOTA Multi-Modal LLMs on document forgery detection: We evaluate the performance of leading multi-modal LLMs against traditional forgery detection methods on the "FindIt Again" \cite{tornes2023receipt} dataset, marking the first comprehensive study of reasoning multi-modal LLMs in this context.
    \item Ablation studies into their reasoning capabilities \& Interpretability: We investigate the reasoning steps employed by multi-modal LLMs, identifying critical features and logical pathways that drive their forgery detection decisions.
    \item Our findings demonstrate that SOTA multi-modal LLMs exhibit robust generalization and competitive performance compared to traditional detection approaches. This study offers valuable insights into the role of reasoning in document forgery detection and establishes a foundation for future research into reliable, interpretable, and generalized detection systems.
\end{itemize}


\section{Related work}

\subsection{Machine learning for Document forgery analysis}

The detection of document forgeries has evolved significantly with the advent of machine learning (ML) and deep learning (DL) techniques. More recently, specialized DL-based methods for document image forgery detection have been proposed, including spatial-frequency and multi-scale feature networks \cite{li2025spatial}, multiscale edge similarity analysis \cite{sun2024multiscale}, and log-transform histogram equalization techniques \cite{bae2025log}. Unsupervised approaches leveraging deep feature clustering have also shown promise for forged document classification \cite{tyagi2024forged}. Additionally, methods exploiting RGB channel inconsistencies \cite{gornale2022rgb} and CNN architectures tailored to administrative documents have achieved strong performance \cite{maamouli2022cnn}. 

\subsection{Datasets for Imagery forgery detection}

While the broader field of image forgery has been more widely investigated, as shown in Table~\ref{tab:forgery_datasets}, the topic of document forgery has gained significantly less traction. To date, only one dataset specifically focused on ID document copy-move forgeries has been introduced \cite{mahfoudi2021cmid}. Given the scarcity of dedicated benchmarks, we evaluate our model performance on the publicly available RDDFD ("Receipt dataset for document forgery detection") dataset. Notably, a similar dataset in a Japanese context was constructed by Okamoto et al., though it is not publicly accessible \cite{okamoto2023deep}.

\subsection{State of the Art Multi-Modal LLMs: Capabilities and Advances}

We list the SOTA LLMs that we benchmark in Table~\ref{tab:SOTA_LLM}, including their MMMU benchmark scores and publication dates. At the time of writing, several of these models’ APIs have only recently become available, enabling novel contributions to this field.

\begin{table}[h!]
\centering
\caption{Public Datasets for Forgery image Detection}
\label{tab:forgery_datasets}
\begin{adjustbox}{width=\textwidth}
\begin{tabularx}{\textwidth}{@{}l c X X  X@{}}
\toprule
\textbf{Dataset} & \textbf{\#images} & \textbf{Image Type} & \textbf{Forgery Type}  \\
\midrule
FID \cite{hossain2022forgery} & 4k & General-purpose images from public repositories & Cut-paste, copy-move, erase-filling  \\
\textbf{CG-1050}\cite{castro2019cg1050} & 1k & Indoor and outdoor scenes (e.g., streets, parks, malls) & Copy-move, cut-paste, retouching, colorizing  \\
\textbf{MIDV-2020}\cite{bulatov2021midv2020} & 74.4k & Synthetic identity documents with unique text fields and artificially generated faces & Simulated identity documents for tasks like text field recognition and face detection  \\
\textbf{ForgeryNet}\cite{he2021forgerynet}  & 3M & Faces and general scenes & Various, including identity-replaced and identity-retained forgeries  \\
\textbf{DF2023} \cite{fischinger2025df2023} & 1M  & Various, including manipulated images from online social networks & Splicing, copy-move, enhancement, removal  \\
\textbf{CASIA v2.0} \cite{Dong2013} & 4.8K & Natural scenes, objects, people & Splicing, copy-move  \\
\textbf{MICC-F220} \cite{amerini2011sift} & 220  & Natural scenes & Copy-move  \\
\textbf{CoMoFoD} \cite{tralic2013comofod} & 260 & Various scenes & Copy-move with transformations (translation, rotation, scaling, distortion)  \\
\textbf{RDDFD}\cite{tornes2023receipt}  & 1K  & Retail receipts & Not explicitly focused on forgery; primarily for OCR and information extraction tasks  \\
\bottomrule
\end{tabularx}
\end{adjustbox}
\end{table}


\begin{table}[h]
    \centering
    \caption{Comparative Vision Performance of Multi-Modal LLMs}
    \label{tab:SOTA_LLM}
    \begin{tabular}{llcc}
        \toprule
        \textbf{Publisher} & \textbf{Model Version} & \textbf{MMMU (\%)} & \textbf{VQA v2 Acc (\%)} \\
        \midrule
        \multirow{2}{*}{OpenAI} & GPT-4o & 69.1 & 85.3 \\
                                & GPT-o1 & 78.2 & N/A \\
        Meta                    & Llama 3.2 & 71.3 & 78.1 \\
        \multirow{2}{*}{Google} & Gemini 2 & 68.5 & 84.5 \\
                                & Gemini 2 Thinking & 75.4 & N/A \\
        \multirow{2}{*}{Alibaba}& Qwen 2 VL & 45.2 & N/A \\
                                & Qwen 2.5 VL & 70.3 & N/A \\
        \multirow{2}{*}{Anthropic} & Claude 3.5 Haiku & 50.2 & N/A \\
                                   & Claude 3.7 Sonnet & 71.8 & N/A \\
        \multirow{2}{*}{Deepseek}  & Janus-Pro 1B & 36.3 &  N/A \\
                                    & Janus-Pro 7B & 41.0 &  N/A \\
        Mistral                 & Pixtral & 52.5 & N/A \\
        \bottomrule
    \end{tabular}
\end{table}

\addtolength{\textheight}{-3cm}   

\section{Methodology}

This study addresses the problem of receipt forgery detection comparing SOTA LLMs with traditional methods like CNN and SVM which provide a baseline as stated in \cite{tornes2023receipt}.


\subsection{SVM-Based Approach}

Each receipt image was preprocessed by resizing to 250$\times$250 pixels to ensure uniform input dimensions. Pixel intensities were normalized using the standard mean and standard deviation values derived from the ImageNet dataset. The metadata files were parsed to assign binary labels (authentic or forged) to each image, ensuring that corrupted or unreadable files were excluded from further processing.

For the SVM-based approach, the resized images were flattened into one-dimensional feature vectors. These vectors were standardized using z-score normalization to ensure consistent scaling across features. A linear-kernel Support Vector Machine was trained on the extracted features, with class weights adjusted to mitigate class imbalance in accordance with the method used in  \cite{tornes2023receipt}.
Further, grid search was used to experiment with different 'kernel', 'gamma', 'c' and 'class weight' parameters to find best model that does not overfit.

\subsection{CNN-Based Approach}

the method is a deep learning approach (CNN) applied on OH-JPEG/OH-JPEG+PQL features for detecting document forgeries mentioned in \cite{tornes2023receipt}. 


The OH-JPEG method utilizes a one-hot encoding of the DCT (Discrete Cosines Transformed) coefficients of JPEG images to capture spatial and frequency patterns in JPEG images. The methods work by transforming each 128x128 pixel block into DCT coefficients, where low frequencies (top-left of DCT block) represent basic structure and high frequencies (bottom-right) represent fine details, allowing the network to learn both obvious manipulations through structural changes and subtle forgeries through compression artifacts.

while OH-JPEG+PQL enhances this by incorporating PQL that multiplies features with the image's quantization matrix to capture compression artifacts and quality inconsistencies.

The Parity Quantization layer (PQL) enhancement makes the system particularly robust for real-world document forensics as it can detect various forgery types, e.g copy and paste inside/outside the document, as mentioned in \cite{tornes2023receipt}) by analyzing both content and compression characteristics, with the quantization information helping identify even subtle manipulations that maintain visual consistency but leave traces in compression patterns. The methods are particularly effective for document forgery detection because they analyze both structural changes (through Discrete Cosine Transform(DCT) coefficients) and compression characteristics (through PQL), making them suitable for detecting copy-paste operations (by identifying repeated DCT patterns), text imitation (through compression inconsistencies), and pixel modifications (via local pattern disruptions).

After extracting features, The CNN-based model utilized a lightweight convolutional neural network architecture composed of sequential convolutional layers with ReLU activations and pooling layers, followed by fully connected layers for binary classification. The network was trained using the cross-entropy loss function optimized with the Adam optimizer. Evaluation was performed at the end of each epoch on the test datasets.


\subsection{LLM Approach}
\begin{figure*}[h!]
    \centering
    \includegraphics[width=0.6\textwidth]{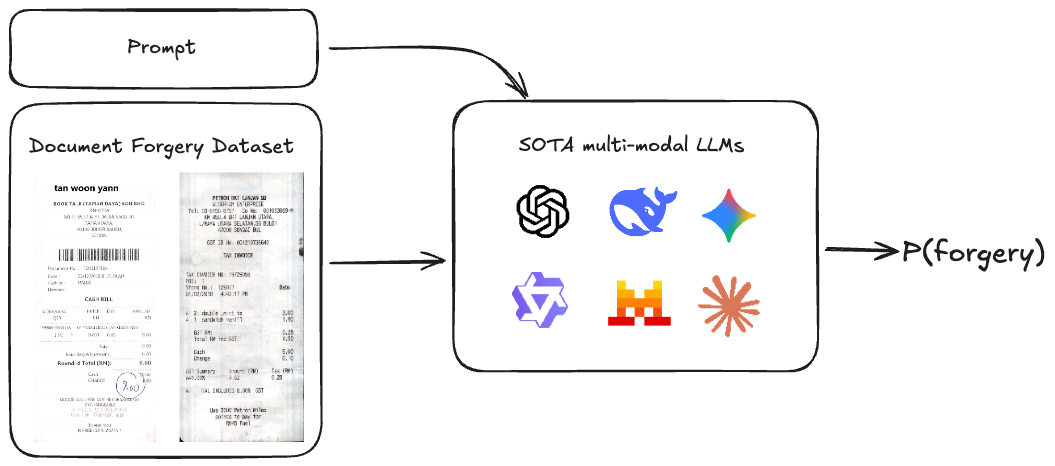}
    \caption{Overall Experiment Design}
    \label{fig:design}
\end{figure*}

This section outlines the methodology employed to investigate the efficacy of state-of-the-art (SOTA) LLMs in detecting document forgery images using a structured pipeline, as depicted in the provided diagram Fig. \ref{fig:design}. The pipeline leverages the find-it-2 datasets, processes them through SOTA multi-modal LLMs with tailored prompts, and generates a probability of an image being forgery, denoted as P(forgery) in a zero-shot manner. 

\subsubsection{ Prompt Engineering and Input to SOTA Multi-Modal LLMs}
The preprocessed images from the fake receipt datasets are fed into a suite of SOTA multi-modal LLMs, which are capable of processing both visual and textual data. The input process is guided by carefully crafted prompts designed to elicit reasoning and classification capabilities from the models. The prompt, as indicated in the diagram, directs the LLMs to analyze the input images and determine their authenticity. After some iterations we landed on the final prompt, shown in Appendix. Each model receives the same prompt, ensuring consistency across evaluations.




\section{Results}

\subsection{ Evaluation and Analysis}

To assess the performance of the pipeline, the P(forgery) scores generated by each SOTA multi-modal LLM are benchmarked against ground truth labels from the document forgery datasets. Since language models does not necessarily returns structured outputs, we process the output with string processors to extract the numeric answer of the model. Note that this process does not guarantee full recovery of the prediction score, combined with a small probability that LLMs refuse to give a numeric estimation or just fails to reply. We omit those examples in our result calculation. 

For the evaluation metrics, we use the ROC curves with area under the receiver operating characteristic curve (AUC).  
Additionally, an in-depth analysis of the reasoning pathways is conducted to identify the key contributing factors. 

To further understand and benchmark whether the LLMs can perform, we benchmarked traditional SVM and Computer Vision models that were in prior publications.


Given the potential class imbalance between authentic and forged receipts, particular attention was given to the AUC score to ensure balanced performance across both classes.

\subsection{SVM and CNN Performance}

On the one hand, SVM performed poorly with  f1 score of \textbf{0.18} without hypertuning. The best model found after gridsearch did some effort of reducing overfitting, and the f1 score for that was found to be \textbf{0.24}. The best model however gives AUC of\textbf{0.48}, which is even worse than random guessing. The first CNN model described in \cite{tornes2023receipt} gave best f1 score of \textbf{0.69} on test dataset , but its confusion matrix shows the model has overfitted and predicted everything as forged. We tried more adjustments like adjusting classification thresholds and increasing weights of minority classes, but the model ended up overfitting and marking all documents as forged. A similar thing happened on using ResNet50 pretrained model, because its architecture becomes too complex for this task.

On the other hand, while the original paper\cite{tornes2023receipt} claimed a 0.79 precision on OH‑JPEG model and 0.78 precision on OH-JPEG/PQL, which we reprodcued(OH-JPEG: 0.80 F1 score, and 0.71 precision, and the OH‑JPEG‑PQL variant similarly delivers strong performance with 0.74 F1, 0.77 precision). These results are on patch level, which a forged image contribute a forged patch and two pristine patches, according to the paper\cite{tornes2023receipt}. To get the performance on image level, which is more intuitive on inference, when aggregate predictions for each image (use OR logic to determine if any crop is forged). At image-level aggregation, although we have matched the reported performance in F1 scores with previous publication, we found the AUROC of the detection performance of such traditional method is actually around 0.5, show casing strong gap in the reporting metric from previous work. 






\subsection{LLM Performances}

In this section, we show the performance results on different LLMs and discuss the following questions. and analysis the performance on GPT-O1 because it has the best performance on this task.

\subsubsection{Is Document Forgery Detection a Simple Task for LLMs?}

\begin{figure}[ht]
  \centering
  \includegraphics[width=0.5\linewidth]{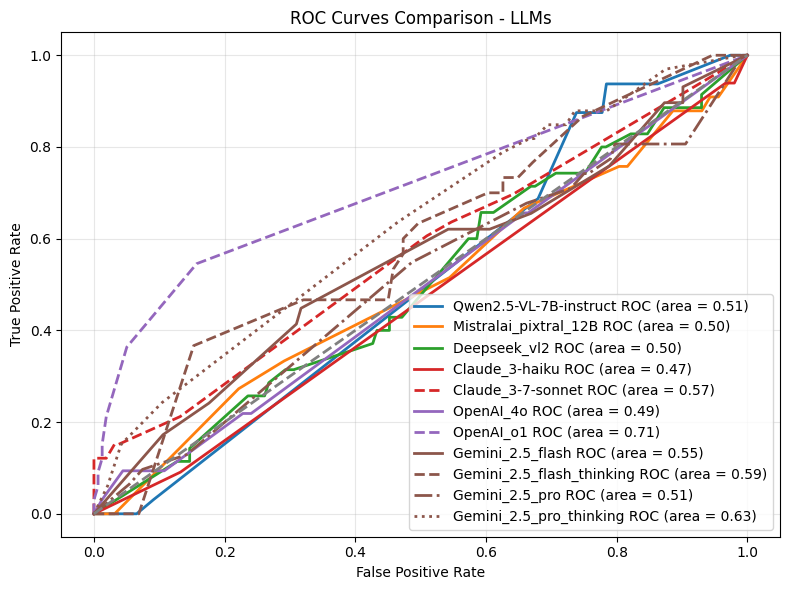}
  \caption{ROC curves showing LLM performance on document forgery detection.}
  \label{fig:roc-curves-llm}
\end{figure}

For most of the LLMs, the AUC scores are close to 0.5, which is a random guess. among which, GPT-O1 is the top performer with a 0.71 AUC score. All ROC curves are shown in Figure\ref{fig:roc-curves-llm} and Table. \ref{tab:auc-summary} summarizes the AUC scores of LLMs.

\begin{table}[ht]
  \centering
  \begin{tabular}{lcc}
    \toprule
    \textbf{Model} & \textbf{AUC} \\

    \midrule
    \textbf{GPT-O1}                         & \textbf{0.71} \\
    GPT-4o                                  & 0.49  \\
    \textbf{Gemini-2.5-Pro-Think}           & \textbf{0.63}  \\
    Gemini-2.5-Flash-Think                  & 0.59 \\
    Llama-4                                 & 0.60 \\
    \textbf{Claude-3.7}                     & \textbf{0.57} \\
    Claude-3-Haiku                          & 0.47  \\
    Qwen2.5-VL-7B-Instruct                  & 0.51  \\
    Pixtral-12B-2409                        & 0.50 \\
    Deepseek-v12                            & 0.50  \\
    \midrule
    SVM                                     & 0.49  \\
    CNN (OH-JPEG)                           & 0.51 \\
    \midrule
    Baseline (random)                       & 0.50 \\
    \bottomrule
  \end{tabular}
  \caption{AUC scores of all models.}
  \label{tab:auc-summary}
\end{table}




\subsubsection{Does Model Improvement Generalize to Document Forgery Detection?}
\label{subsec:general-improvement}

We conducted a comparative analysis between the GPT series and Claude series models. Our findings demonstrate that the latest model releases exhibit superior performance in detecting document forgeries compared to their predecessors. The results presented in Table. \ref{tab:auc-summary} 



\subsubsection{Does Model Size Correlate with Performance?}
\label{subsec:mid-size}

We conducted an investigation into three mid-sized LLMs (<12B) on the Document forgery detection task. Our findings demonstrate that Document forgery detection presents a formidable challenge for widely adopted mid-sized language models (LLMs). Notably, all AUC scores obtained were approximately 0.5. The results presented in Table. \ref{tab:auc-summary}








\subsubsection{Do Reasoning Models Perform Better?}
\label{subsec:reasoning-performance}

We conducted an experiment on the latest Gemini-2.5 model series. This version of Gemini enables the control of reasoning through a reasoning-token-length variable. In the non-thinking experiment, we explicitly set the reasoning-token-length to zero. Our findings demonstrate that reasoning enhances the model’s performance on this task, as presented in Table. \ref{tab:gemini-thinking-comparison} 


\begin{table}[ht]
  \centering
  \begin{tabular}{lcc}
    \toprule
    \textbf{Model}               & \textbf{w/o Thinking} & \textbf{w/ Thinking} \\
    \midrule
    Gemini-2.5-Flash             & 0.55                      & 0.59                   \\
    Gemini-2.5-Pro               & 0.51                      & 0.63                   \\
    \bottomrule
  \end{tabular}
  \caption{AUC scores for Gemini-2.5 variants w/ and w/o thinking.}
  \label{tab:gemini-thinking-comparison}
\end{table}



\subsubsection{Performance Analysis of GPT-O1}

GPT-O1 is the top performer among all LLMs on this test. This part details the performance analysis of GPT-O1 evaluated in this study, focusing on its reasoning capabilities, and specific strengths and weaknesses in detecting document fraud within the benchmark dataset (218 receipt images compared against \texttt{test.txt} ground truth).

GPT‑O1 demonstrated a moderate overall performance, achieving an accuracy rate of \textbf{83.85\%}, correctly classifying 161 out of 192 samples (31 incorrect determinations). The model showcased impressive capabilities in:
\begin{itemize}
\item \textbf{Mathematical Verification:} Excelling at detecting incorrect line totals, subtotals, flawed Goods and Services Tax (GST) calculations, and incorrect change amounts.
\item \textbf{Contextual and Temporal Logic:} Successfully identifying inconsistencies related to dates (e.g., transaction time vs. closing time) and historically accurate tax rates (e.g., applying correct GST percentages based on the transaction date).
\item \textbf{Content Consistency:} Reliably flagging conflicting information within the document, such as discrepancies between merchant names or addresses in the header versus the footer.
\item \textbf{Visual Anomaly Detection:} Identifying clearly suspicious visual elements like deliberately obscured values or significant font mismatches in critical areas.
\item \textbf{Detailed Reasoning:} Providing clear, specific, and logically structured justifications for its determinations.
\end{itemize}

\begin{figure}[ht]
  \centering
  \includegraphics[width=0.5\textwidth]{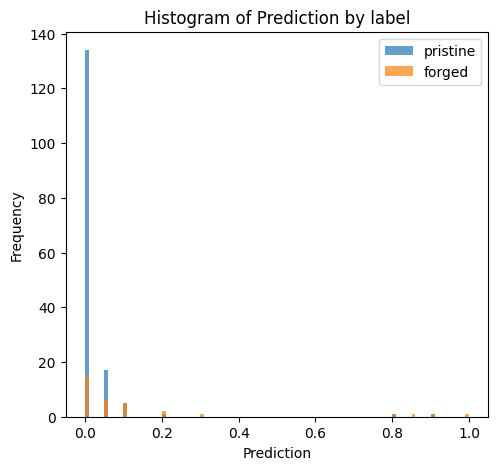} 
  \caption{Histogram of predicted probabilities by true label.}
  \label{fig:histogram_o1}
\end{figure}

Confidence calibration revealed that GPT‑O1 expresses its predictions with generally low probability scores, indicating cautious but underconfident behavior (see Figure~\ref{fig:histogram_o1}). True Negatives (pristine) overwhelmingly received scores clustered around 0.02–0.05, and True Positives (forged) were similarly distributed at 0.05–0.15. Only a handful of cases—namely the 2 false positives and a few true positives—spiked above 0.8, accounting for the rare overconfident peaks in the histogram. Critically, the 29 false negatives lie almost entirely in the lowest bin (<0.05), demonstrating extreme underconfidence when missing forged samples. This mismatch between score magnitude and correctness highlights calibration issues: GPT‑O1 seldom expresses strong confidence even for correct predictions, while its infrequent overconfident errors could mislead any fixed‐threshold decision rule.

The primary weaknesses leading to the 31 errors were:
\begin{itemize}
\item \textbf{False Positives (2 cases):} The most common error type. The model tended to flag legitimate receipts as fraudulent primarily due to conflicting merchant information (e.g., HQ address vs. branch location) which the ground truth considered acceptable, or minor calculation/rounding differences deemed suspicious by the model but legitimate according to the label.
\item \textbf{False Negatives (29 cases):} Less frequent but notable. The model occasionally missed subtle calculation errors or showed over-reliance on internal mathematical consistency, potentially overlooking other red flags if the (flawed) numbers added up correctly.
\end{itemize}
Overall, GPT‑O1 displayed advanced analytical capabilities suitable for detailed document verification, with its main deviation from the ground truth stemming from a stricter interpretation of certain inconsistencies like header/footer discrepancies.

\subsubsection{Do Models Hallucinate on Forgery Detection?}
\label{subsec:hallucinations}

In our experiments, all LLMs demonstrated difficulties in differentiating fraudulent document images from legitimate document images. Details (histogram figures) are shown in the Appendix. The histogramgram figure shows that even the top-performing GPT-o1 model cannot classify fraudulent images effectively. These findings suggest the inherent difficulty of the task and the necessity of employing a specially trained model for this purpose.

\subsection{LLMs compared with traditional computer vision solutions}
By comparing the performance of traditional models, SVM and CNN, that were trained on the tasks with zero-shot performance with the top performing LLMs. We can see that both conventional models and the best LLMs are struggling on this document forgery task. under zero-shot settings (we hypothesize given document forgery detection is too niche, LLMs are very unlikely to have been trained specifically on this task), the best LLMs show limited ability to classify document forgery, and the traditional models consistently performs around random guessing.

\section{Conclusions}
This study demonstrates that state-of-the-art multi-modal large language models (LLMs) with reasoning capabilities offer moderate potential for document forgery detection, achieving competitive performance and notable generalization compared to traditional methods. Our benchmarking reveals that while model size can enhance performance, as seen with GPT series consistent improvements, newer model versions and reasoning-augmented approaches translate to limited outcomes in this niche task (e.g Gemin-2.5-Flash and Pro), with some models (e.g.Qwen, Pixtral and reasoning variants) performing near or below random guessing. OpenAI’s LLMs stood out, surpassing traditional svm and computer vision models in zero-shot generalization, though they remain outperformed by specialized networks on datasets these were trained on. Additionally, our analysis suggests that poor performance is not solely attributable to numerical output limitations, as categorical prompting yielded similar results. These findings underscore the value of integrating multi-modal reasoning into deepfake detection frameworks while highlighting the need for further exploration into failure modes and model hallucination to enhance robustness and interpretability in real-world applications.

\section*{Data and Code Availability}
The datasets and code supporting this work are released by Scam.AI and are available from
the Scam.AI Research portal, \url{https://scam.ai/research}~\cite{scamai_research}. Document-
and PDF-forgery resources are additionally maintained at Trustable, \url{https://trustable.ai}~\cite{trustable_ai}.
These are the canonical, maintained locations for the resources associated with this paper and
supersede any prior ad hoc distribution.

\bibliographystyle{ieeetr}
\bibliography{references}

\clearpage
\appendix
\section{Prompt Used for Model Experiments}

Below is the exact prompt we supplied to the model in our experiments:

\begin{verbatim}
Prompt:


Task: Receipt Fraud Detection Analysis

Examine the provided receipt image carefully and perform a comprehensive fraud detection analysis. Determine if this receipt is legitimate or fraudulent.

Step 1: Systematic Visual Assessment
- Analyze the document layout, including alignment, spacing, and formatting consistency
- Examine text quality (font consistency, printing quality, resolution)
- Check for visual artifacts (smudges, unusual erasures, digital manipulation signs)
- Assess color consistency and transitions across the document

Step 2: Content Verification
- Verify date and time formatting consistency and logical sequence
- Examine merchant information (name, address, contact details) for completeness and consistency
- Analyze item descriptions, quantities, and pricing for logical coherence
- Verify mathematical accuracy (subtotals, taxes, discounts, final amounts)
- Check transaction identifiers (receipt numbers, transaction IDs) for proper formatting

Step 3: Contextual Evaluation
- Assess logical relationships between items purchased
- Check for reasonable pricing in relation to items purchased
- Verify time of purchase is appropriate for the merchant type
- Evaluate authenticity markers (barcodes, QR codes, watermarks, security features)

Step 4: Potential Fraud Indicators
- Identify inconsistencies in numerical values or calculations
- Detect misaligned text or irregular spacing patterns
- Note any unusual edits, erasures, or modifications
- Flag unrealistic purchases, prices, or merchant information

Conclusion:
Based on your analysis, provide a decisive determination of the question: Is this receipt fraudulent? 
1. confidence level 
2. List specific evidence supporting your determination.
3. Identify any remaining uncertainties or areas requiring further investigation.

State your confidence level from 0 to 1, 0 for not confident at all(it's legitimate), 1 for certainty (it's fraudulent).

Give your conclusion in json format:
{
    "confidence": float,
    "evidence": string,
    "remaining uncertainiies": string.
}
\end{verbatim}

\end{document}